\pdfoutput=1

\documentclass[11pt]{article}

\usepackage[final]{acl}

\usepackage{times}
\usepackage{latexsym}

\usepackage[T1]{fontenc}

\usepackage[utf8]{inputenc}

\usepackage{microtype}

\usepackage{inconsolata}

\usepackage{graphicx}

\usepackage{bm}
\usepackage{subfig}
\usepackage{amsmath}
\usepackage{booktabs}
\usepackage{multirow}
\usepackage{graphicx}
\usepackage{amssymb}
\usepackage{amsmath}
\usepackage{amsthm}
\usepackage{colortbl}
\usepackage{algorithm}
\usepackage{algorithmic}
\usepackage{subfig}
\title{DV-SFT: Direct Vision Supervision for Fine-Grained Visual Understanding}

\author{
 \textbf{Jianfei Zhao\textsuperscript{1,2}},
 \textbf{Feng Zhang\textsuperscript{1}},
 \textbf{Xin Sun\textsuperscript{1}},
 \textbf{Chong Feng\textsuperscript{1,5}},
 \textbf{Bing Wang\textsuperscript{3}},
 \textbf{Zhixing Tan\textsuperscript{4}}
\\
 \textsuperscript{1}School of Computer Science and Technology, Beijing Institute of Technology, \\
 \textsuperscript{2}Zhongguancun Academy,
 \textsuperscript{3}Beihang University, 
 \textsuperscript{4}Zhongguancun Laboratory \\
 \textsuperscript{5}Southeast Academy of Information Technology, Beijing Institute of Technology, \\
\small{
 {zhqingan}{@bit.edu.cn}
 }
}

\begin{document}
\maketitle
\begin{abstract}

Multimodal large language models are typically trained end-to-end to predict ground-truth answers, yet supervision signals are applied exclusively to text tokens. Visual tokens, the core carriers of visual information, are optimized only implicitly as part of the context, leading to coarse-grained visual understanding.
Prior works attempt to supervise visual inputs but inevitably rely on auxiliary components such as additional decoders or forward passes, because visual tokens lack readily interpretable labels. This limits their practical applicability.
In this work, we propose \textbf{D}irect \textbf{V}ision \textbf{S}upervised \textbf{F}ine-\textbf{T}uning (DV-SFT), which constructs explicit, token-level supervision for visual tokens and trains them through the same next-token prediction objective used for text.
Specifically, we exploit the direct vision--text correspondence in OCR-related scenarios and automatically label each visual token with the word in its corresponding image patch.
DV-SFT treats the MLLM as a black box, requiring no architectural modifications or additional forward passes.
Extensive experiments demonstrate the superiority of direct vision supervision. DV-SFT consistently outperforms standard SFT across three in-domain and four out-of-domain benchmarks.
Further analyses show that vision supervision effectively enhances fine-grained visual understanding and achieves higher multimodal alignment efficiency.
\footnote{The code will be released once accepted.}

\end{abstract}

\section{Introduction}

Multimodal large language models (MLLMs) \citep{llava, BLIP, qwen3-vl} augment large language models with a vision encoder \cite{ViT, SigLIP}, mapping images into the language feature space. End-to-end training aligns multimodal features and enables MLLMs to achieve near-human performance on various visual understanding tasks. 
Despite this success, a critical limitation remains: all explicit supervision signals are applied solely to text tokens. Visual tokens---the primary carriers of visual information---are only optimized implicitly as part of the context, receiving no direct semantic guidance.
Scaling up end-to-end training, though beneficial in the aggregate, still supervises visual tokens only implicitly through data statistics, leaving fine-grained details easily overlooked or hallucinated \citep{VS, BASIC}.

To improve the precision of visual understanding, a number of recent efforts attempt to inject additional supervision for visual tokens.
However, these signals are often indirect and coarse-grained, such as devising vision-targeted auxiliary rewards \citep{Vision-SR1, DyME} or distilling visual features through representation alignment \citep{BASIC, LaVer, PatchAligned}. 
In other cases, the training objective of visual tokens is misaligned with visual understanding, as seen in next-visual-token prediction \citep{ASVR, emu3, EchoGen} or pixel-level reconstruction \cite{ROSS, DS-VLM}. 
More critically, most of these methods require architectural modifications such as additional decoders or forward passes, which hinder their practical adoption.
A more direct, intuitive, and architecturally simple form of vision supervision thus remains an open need.

An intriguing observation from recent studies \citep{VGA, BASIC, IC} suggests a more direct path: end-to-end trained MLLMs can extract semantic information from visual tokens without any dedicated vision supervision. Specifically, for each visual token, the words corresponding to peak values in its logits (hereafter \emph{visual logits}) often accurately describe the content of the associated image patch, including objects, attributes, text strings, and so on. 
Inspired by this finding, we hypothesize that explicitly supervising visual tokens by the words that describe their corresponding image patches can substantially enhance fine-grained visual understanding.

\begin{figure}
    \centering
    \includegraphics[width=\linewidth]{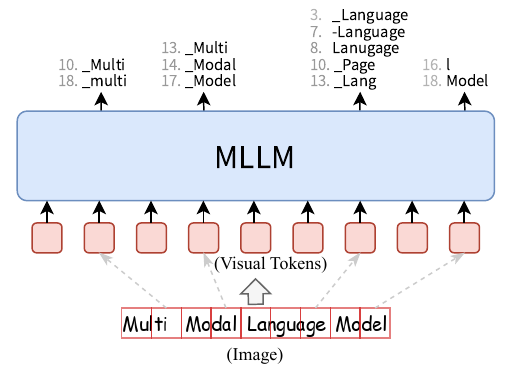}
    \caption{An example of visual logits in Qwen3-VL-2B showing the top-20 tokens at the final image patch of each word. Numbers indicate rank; punctuation and meaningless characters are removed.
    This observation inspires us to \emph{supervise visual tokens using the words in the image patches}.
    }
    \label{fig:case}
\end{figure}

To validate this idea, we concentrate on OCR-related scenarios, where the correspondence between vision and text is most straightforward, facilitating fine-grained alignment between image tokens and words.
As illustrated in Figure~\ref{fig:case}, an MLLM's visual logits in an OCR image can roughly recognize the textual content, but the granularity remains coarse and the alignments imprecise---a direct consequence of the absence of fine-grained vision supervision.
The OCR scenarios thus serve as a clean, well-controlled testbed for studying direct vision supervision.

We therefore propose \textbf{D}irect \textbf{V}ision \textbf{S}upervised \textbf{F}ine-\textbf{T}uning (DV-SFT), a simple and principled method that uses words in visual tokens as vision labels and applies the next-token prediction loss directly on visual tokens. Critically, DV-SFT requires no architectural modifications, no additional forward passes, and treats the MLLM as a black box, making it readily applicable to any existing MLLM pipeline.
We conduct comprehensive evaluations across eight benchmarks, including document and chart understanding \citep{squad, docvqa, infovqa, chartqa}, OCR \citep{ocrvqa, ocrbench, textvqa}, and general VQA~\citep{mme}.

In summary, our main contributions are as follows:
\begin{itemize}
\item We propose DV-SFT, which enables direct token-level supervision on visual tokens without any modification to model architecture or forward path, keeping a fully unified, end-to-end training paradigm.
\item We design an offline vision label construction pipeline that leverages OCR and drawing tools to produce fine-grained, token-level supervision for visual tokens.
\item Extensive experiments show that DV-SFT significantly improves visual capability and outperforms standard SFT across a range of benchmarks, with detailed analyses confirming that the gains stem from refined visual representations.
\end{itemize}

\section{Related Work}

Multimodal large language models \citep{llava, qwen3-vl} typically adopt an end-to-end training paradigm, where visual tokens receive supervision only indirectly through the correctness of the final text output. Although this approach effectively models visual features, the learning outcomes are often unstable, leading to coarse-grained interpretation and hallucination \cite{HalluSurvey, VS, CICD}.
To mitigate this, a growing body of research has explored incorporating supervision on the visual tokens.

Existing vision supervision methods can be broadly grouped into four paradigms.
(1) The first provides \emph{indirect semantic guidance} by using reward signals.
Reward functions offer a more flexible form of supervision, and these methods \citep{Vision-SR1, DyME} reward the model based on the visual consistency observed in sampled outputs.
While these approaches effectively strengthen visual understanding, they remain an indirect form of supervision.
(2) The second paradigm employs \emph{representation alignment}.
These methods \citep{BASIC, LaVer, PatchAligned} distill high-quality features to supervise the forward pass of visual tokens.
However, the poor interpretability of continuous representations can inject noise and limit the reliability of the supervision.
(3) The third paradigm is \emph{next-visual-token prediction}.
The emergence of discrete visual tokens has bridged the gap between the learning forms of visual and text tokens. Methods in this category supervise visual tokens through autoregressive image generation tasks \citep{emu3, EchoGen}.
Despite their explicitness, these approaches suffer from a misalignment between the generative objective and the goal of visual understanding
\citet{ASVR} retain the architecture of visual understanding models and use only discrete visual tokens as supervisory signals, but the objective misalignment remains.
(4) The fourth paradigm aims for \emph{reconstructing visual input}.
These studies \citep{ROSS, DS-VLM} supervise the correctness of visual token understanding by reconstructing the original visual input from visual representation; however, they face a conflict \citep{DualToken} between pixel-level and semantic features, resulting in suboptimal learning outcomes.
Importantly, nearly all of these methods introduce additional architectural components, such as visual tokenizers, diffusion-based decoders, or extra forward passes, which complicate training and limit their general applicability.

In contrast, our proposed DV-SFT avoids all of the above limitations.
It constructs direct token-level labels for visual tokens based on the visual semantics in the image patches, closely aligning with the goal of visual understanding.
By modeling vision supervision as a standard next-token prediction task, DV-SFT achieves a fully unified, end-to-end training paradigm.
More importantly, DV-SFT introduces no additional modules or forward propagation passes, offering high practical usability.

\section{Preliminaries}
An MLLM typically consists of a vision encoder that processes visual inputs, a projector that aligns features across modalities, and a large language model.
In MLLMs' training process, the input image is first tokenized into patches, and the patch sequence is input into the visual encoder.
After successively processed by the encoder and projector, the input image finally forms a sequence of visual embeddings $\bm{I}=[\bm{v}_1,\bm{v}_2,\cdots,\bm{v}_m] \in \mathbb{R}^{m \times D}$ where $m$ is the number of visual tokens and $D$ is the model dimension.
When concatenated with the text input embeddings, the entire sequence is fed into the large language model, which outputs the logits for each token.
The language modeling loss will act on the text logits under the teacher-forcing paradigm:
\begin{equation}
    \mathcal{L}_t = -\frac{1}{n}\sum^n_i \log \mathrm{P}(y_i|\bm{I},\bm{X},\bm{Y}_{<i}; \theta),
\end{equation}
where $\bm{X}$ is the prompt and $\bm{Y}$ of length $n$ is the ground-truth response.

\begin{figure}
    \centering
    \includegraphics[width=0.9\linewidth]{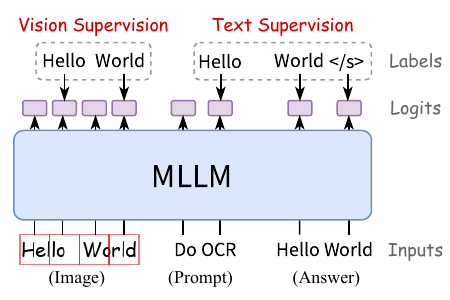}
    \caption{Diagram of DV-SFT.
    The training procedure of DV-SFT is identical to that of standard SFT, where all tokens are trained end-to-end on the next-token prediction task. The difference lies in the fact that a significant portion of visual tokens have valid labels.
    }
    \label{fig:diagram}
\end{figure}

Like text tokens, the model also produces logits for all visual tokens, referred to as visual logits.
Recent studies \citep{VGA, IC, BASIC} found that the distribution of visual logits contains concrete semantic features in each visual token.
Sepcifically, the top words in $\mathrm{P}(.|\bm{I}_{\le i})$ always reflect the visual content in the patch of $\bm{v}_i$.
Intuitively, the generalizable feature patterns that emerge spontaneously in models under end-to-end training serve as high-quality supervisory signals.
Therefore, inspired by this, we use words that are semantically aligned with the visual content to supervise the learning of visual tokens in MLLMs:
\begin{equation}\label{Eq_DV}
    \mathcal{L}_v = -\frac{1}{m}\sum^m_i \log \mathrm{P}(w_i\mid \bm{I}_{\le i}; \theta),
\end{equation}
where $w_i$ is the label for the visual token $\bm{v}_i$.

\section{Direct Vision Supervision}

We propose Direct Vision Supervised Fine-Tuning (DV-SFT), which utilizes words that describe image patches as direct supervision signals for visual tokens, thereby enhancing the model's learning of visual tokens.
Due to the information-rich nature of the visual modality, constructing labels for visual tokens is highly challenging.
To address this, we exploit the straightforward association between vision and text in OCR-related scenarios to construct labels for visual tokens.
Specifically, we use the word contained in a visual token as its label and let the model learn through the next-token prediction task, as illustrated in Figure~\ref{fig:diagram}.
To ensure parallel training, we perform only one-step prediction for vision labels. Specifically, the loss is computed using only the first token after tokenizing the vision label. This practice also conforms to the first-token preference property of visual logits \citep{VGA}.

\subsection{Vision Label Construction}

We attempt to construct labels for visual tokens in OCR-related scenarios, where the direct correspondence between text and vision makes the construction of vision labels feasible.
The challenge in constructing vision labels lies in establishing a one-to-one mapping between words in the image and visual tokens. Specifically, it involves identifying which words appear in the image and determining which visual tokens each word should supervise.

We achieve this through two approaches: (1) Image-to-Label, and (2) Label-to-Image.

\paragraph{Image-to-Label.}
We first use a text detection model and a text recognition model to conduct word-level OCR for the image, obtaining all words along with their spatial locations.
We identify word concatenation errors in the OCR results by checking the tokenization length and discard words whose tokenization length exceeds three.
Extremely large words (with a height exceeding three times the patch size) are also discarded to ensure the annotation precision of the vision labels.
Considering the autoregressive nature of MLLMs, we align each word to the visual token corresponding to its bottom-right corner and use it as the label for that visual token.
Finally, we filter out multiple vision labels assigned to the same visual token to avoid confusing the model.

\paragraph{Label-to-Image.}
The visual encoder tokenizes images through a mechanical strategy, which prevents perfect alignment between text in the image and visual tokens.
To address this issue, we construct fine-grained vision labels through a text-to-image approach.
Specifically, we use a unimodal document question-answering dataset and employ drawing tools to render documents by aligning each word individually onto the patch grid of a blank image.
Words that span multiple visual tokens are assigned as labels only to the last visual token, in accordance with the autoregressive property.

\subsection{Vision Smoothing}

Although MLLMs are causal models with unidirectional attention, the visual encoder possesses bidirectional characteristics. This means that visual information will interact across tokens—that is, a visual token may contain semantic information from subsequent visual tokens in the sequence.
The example in Figure~\ref{fig:case} illustrates this phenomenon: the last word ``model'' in the image appears in the logits of the fourth token.

To address this phenomenon, we propose Vision Smoothing method inspired by label smoothing, in which a visual token learns its own vision label while also learning other vision labels in the image.
Accordingly, Eq.~\ref{Eq_DV} is modified as:
\begin{equation}\label{Eq_vsm}
\begin{split}
\mathcal{L}_v = &-\frac{1}{m}\sum_{i=1}^m 
  \Bigg[ (1-\beta) \log \mathrm{P}(w_i\mid \bm{I}_{\le i}; \theta) \\
  &+ \frac{\beta}{|V|-1} \sum^V_{t \neq w_i} \log \mathrm{P}(t\mid \bm{I}_{\le i}; \theta) \Bigg],
\end{split}
\end{equation}
where $V$ denotes all vision labels in an image and $\beta$ is the smoothing factor.

The final loss of DV-SFT is:
\begin{equation}\label{Eq_lambda}
\mathcal{L} = \mathcal{L}_t + \lambda \mathcal{L}_v,
\end{equation}
where $\lambda$ is a hyperparameter.

\section{Experiments}

\subsection{Experimental Setup}

\begin{table*}[]
    \setlength{\tabcolsep}{3mm}
    \centering
    \begin{tabular}{l|ccccc}
    \toprule
    Dataset & Samples & Images & Text Labels & Vision Labels & Vision Coverage \\
     \midrule
    DocVQA & 39,463  &  10,194 &  190,895 &  5,402,739 & 6.97\% \\
    InfographicVQA  & 23,946  & 4,406 &  61,608 & 4,173,608 & 11.94\%  \\
    SQuAD   &  24,929 &  13,183  & 60,957  &  2,607,031 &  31.44\% \\
     \bottomrule
    \end{tabular}
    \caption{
    Statistical information of training data. Visual Coverage refers to the percentage of visual tokens that are assigned vision labels.
    }
    \label{tab:data_info}
\end{table*}

\paragraph{Datasets.}
We employ DocVQA \citep{docvqa}, InfographicVQA \citep{infovqa}, and SQuAD \citep{squad} to build training data.
Both DocVQA and InfographicVQA are multimodal datasets, and we construct vision labels using the data-to-label approach.
SQuAD is a unimodal document understanding dataset with extractive answers, which aligns well with the model's need for accurate visual perception.
We convert this dataset into a multimodal task through the label-to-data approach.
The statistics of the training data are shown in Table~\ref{tab:data_info}.

\paragraph{Benchmarks.}
For in-domain data, we use the test sets of DocVQA and InfographicVQA, as well as the validation set of SQuAD. For out-of-domain data, we use the test set of ChartQA \citep{chartqa}, OCRBench \citep{ocrbench}, the test set of OCRVQA \citep{ocrvqa}, and the validation set of TextVQA \citep{textvqa}.
For OCRVQA, we randomly select 3,000 test samples and ensure that the ground truth of each sample is visible in the image.

\paragraph{Baselines.}
We choose the standard SFT method as our baseline, which trains the model using only text labels.
In addition, we reproduce BASIC \citep{BASIC} method, which is a self-distilled representation supervision method.

\paragraph{Implementation Details.}
We select Qwen3-VL-2B/8B-Instruct and Qwen3-1.7B as the base models.
For Qwen3-VL, we randomly select one QA pair (two for InfographicVQA) per image to avoid duplicate vision labels. For Qwen3, we load the visual encoder from Qwen3-VL-2B and use all training data.
We set $\beta=0.3$ in Eq. \ref{Eq_vsm} and $\lambda=2\text{e-}3$ in Eq. \ref{Eq_lambda}. We employ greedy decoding during inference with the default setting.

Detailed experimental setups are provided in Appendix~\ref{sec:detailed setup}.

\begin{table*}[]
    \setlength{\tabcolsep}{1.5mm}
    \centering
    \begin{tabular}{l|cccc|ccccc}
    \toprule
    \multirow{2}{*}{Method} & \multicolumn{4}{c|}{In Domain} & \multicolumn{5}{c}{Out of Domain} \\
    \cline{2-5}\cline{6-10}
     & \small{DocVQA} & \small{InfoVQA} & \small{SQuAD} & \small{Avg.} &  \small{ChartQA} & \small{OCRBench} & \small{OCRVQA} & \small{TextVQA} & \small{Avg.} \\
     \midrule
     \emph{\small{Qwen3-VL-2B}}  &  92.25 & 70.18 & 77.43 & 79.95 & \textbf{70.56} & 72.00 & 72.97 & \textbf{84.88} & 75.10  \\
     SFT   &    93.54 & 71.66 & 86.99 & 84.06 & 69.60 & 73.10 & 75.00 & 84.10 & 75.45     \\
     BASIC  &    84.02 & 57.15 & 72.95 & 71.37 & 54.96 & 65.90 & 66.33 & 77.12 & 66.08    \\
     \rowcolor{gray!20}
     DV-SFT  & \textbf{93.68} & \textbf{72.34} & \textbf{87.38} & \textbf{84.47} & 69.80 & \textbf{74.40} & \textbf{75.17} & 84.08 & \textbf{75.86}    \\
     \midrule
     \emph{\small{Qwen3-VL-8B}}  &  95.97 & 81.88  & 80.27 & 86.23 & 77.12 & 82.30 & 76.03 & 87.80 & 80.81  \\
     SFT   &    96.28 & 82.24 & 87.60 & 88.71 & 77.52 & 83.70 & \textbf{77.17} & 88.44 & 81.71     \\
     BASIC  &     96.38 & 82.39 & 86.72 & 88.50 & 77.44 & 83.40 & 77.00 & 88.14 & 81.50   \\
     \rowcolor{gray!20}
     DV-SFT  & \textbf{96.40} & \textbf{82.49} & \textbf{88.14} & \textbf{89.01} & \textbf{77.76} & \textbf{84.00} & \textbf{77.17} & \textbf{88.48} & \textbf{81.85}  \\
     \midrule
     \emph{\small{Qwen3-1.7B}} & - & - & - & - & - & - & - & - & - \\
     SFT   &    86.49 & 58.02 & 77.60 & 74.04 & 48.60 & 57.90 & 60.67 & 66.74 & 58.48     \\
     \rowcolor{gray!20}
     DV-SFT  &   \textbf{86.57} & \textbf{58.51} & \textbf{78.69} & \textbf{74.59} & \textbf{49.08} & \textbf{58.60} & \textbf{62.20} & \textbf{67.82} & \textbf{59.43}    \\
     \bottomrule
    \end{tabular}
    \caption{
    Main results. \emph{Avg.} denotes the average accuracy. The \textbf{bolded} results indicate the best performance in each setting.
    }
    \label{tab:main_res}
\end{table*}

\subsection{Main Results}
We validate the superiority of our method on both in-domain and out-of-domain data, and the results are illustrated in Table~\ref{tab:main_res}.

\paragraph{In Domain.}
Our method, DV-SFT, achieves consistent advantages on in-domain data, and the experimental results across three base models fully validate its effectiveness. Moreover, the next-token prediction learning paradigm adopted by DV-SFT aligns more naturally with the inherent characteristics of language models, yielding significant advantages over BASIC.
The intuitively comprehensible supervisory signal is another strength of DV-SFT. The representation supervision approach employed by BASIC can suffer from severe model degradation when the quality of the representations is suboptimal. This issue is confirmed in the experiments on the 2B model, where the training results of BASIC fall far behind the original capabilities of the base model. In the experiments on the 8B model, BASIC benefits from higher-quality supervised representations and achieves stable performance improvements; however, its gains are not significantly superior to those of SFT.
Thanks to direct vision supervision, DV-SFT enables the model to understand visual features more accurately, thereby achieving a significant advantage in text-oriented visual understanding.

\paragraph{Out of Domain.}
DV-SFT also demonstrates significant advantages on out-of-domain tasks, indicating its strong domain generalization capability.
Due to the limited knowledge capacity of the 2B model, DV-SFT achieves the best overall performance but does not consistently outperform across all datasets. In contrast, in the experiments with the 8B model, where the larger parameter count affords greater knowledge capacity, DV-SFT consistently improves performance on out-of-domain datasets.
Experimental results on out-of-domain tasks indicate that DV-SFT does not simply improve task-specific performance but genuinely enhances the model's visual capability.

In summary, constructing vision supervision for MLLMs can more effectively enhance their visual understanding capability. Compared with SFT, DV-SFT achieves consistent advantages on both in-domain and out-of-domain tasks. Moreover, the direct vision supervision established by DV-SFT aligns perfectly with the next-token prediction learning paradigm of the model and does not rely on any additional network modules or forward passes. The end-to-end black box training approach requires no intermediate states of the model. These advantages in the training paradigm make DV-SFT highly practical and readily applicable.

\subsection{Multimodal Alignment}
\begin{figure}
    \centering
    \includegraphics[width=\linewidth]{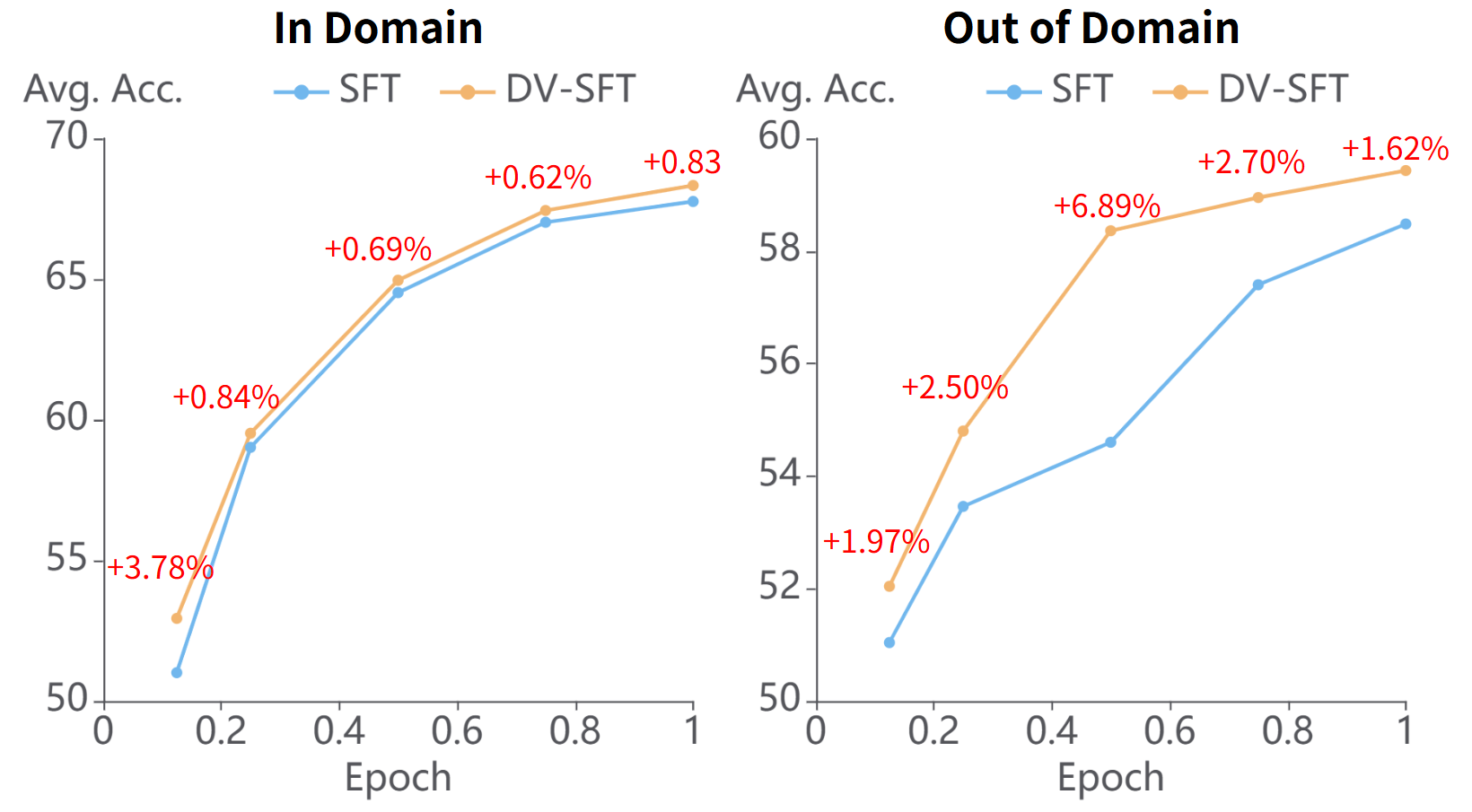}
    \caption{
    Test results of Qwen3-1.7B at each checkpoint during training.
    Results under \emph{In-Domain} are obtained on the validation set.
    }
    \label{fig:alignment}
\end{figure}

To investigate the advantages of vision supervision over text-only supervision, we analyze the effectiveness of the two methods in the process of a unimodal model learning visual features from scratch.
As shown in Table~\ref{tab:main_res}, vision supervision demonstrates comprehensive advantages over text-only supervision, indicating that with limited training data, vision supervision enables the model to better understand visual information.
The advantage of vision supervision is even more pronounced on out-of-domain tasks. Compared with an average performance improvement of 0.74\% on in-domain tasks, DV-SFT achieves an average improvement of 1.6\% on out-of-domain tasks. This suggests that vision supervision genuinely enhances the model's visual understanding capability, rather than merely improving task-specific learning.

We further analyze the visual learning process of the unimodal model, as shown in Figure~\ref{fig:alignment}. It can be observed that in the early stages of training, vision supervision exhibits a more pronounced advantage over text-only supervision. Similarly, this advantage is more significant on out-of-domain tasks.
This indicates that applying direct supervision to visual tokens leads to higher multimodal alignment efficiency compared to indirect supervision of text-only labels.

In summary, the experimental results on the unimodal model further validate the superiority of direct vision supervision.

\subsection{Character Recognition Ability}
\begin{figure*}
    \centering
    \includegraphics[width=0.9\linewidth]{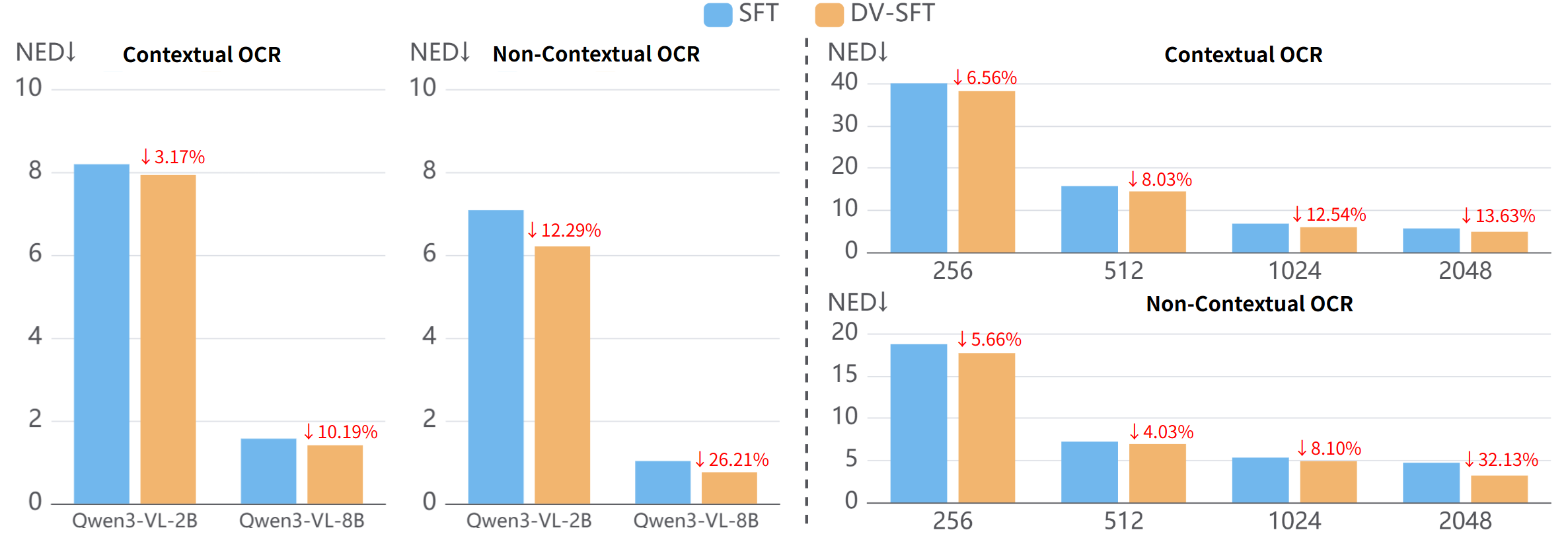}
    \caption{Test results on naive text OCR tasks. \emph{Contextual OCR} refers to the recognition task on normal text, while \emph{Non-Contextual OCR} refers to the recognition task on unordered word sequences.
    \emph{NED} denotes Normalized Edit Distance, where lower values indicate better performance.
    \emph{Left part}: test results at the original image resolution. \emph{Right part}: test results of the 2B model at different image resolutions, with the X-axis representing the visual token length.
    }
    \label{fig:ocr}
\end{figure*}
We use a plain text OCR task to evaluate the improvement in character recognition ability brought by DV-SFT. Specifically, we construct test samples from the news dataset XSUM \citep{xsum}, convert the documents into images using drawing tools, and then ask the model to recognize the text in the images. To avoid the influence of the model's prior knowledge on character recognition, we further construct a Non-Contextual text recognition task. We collect all words from the dataset and sample them based on word frequency to form semantically meaningless word sequences, which are used to test the model's pure character recognition ability.

The results on both tasks, each with 500 test samples, are shown in the left part of Figure~\ref{fig:ocr}.
First, we can observe that DV-SFT achieves superior performance on both OCR tasks, indicating that vision supervision can effectively improve the model's character recognition accuracy. Furthermore, the advantage of DV-SFT is amplified on the Non-Contextual OCR task, which further demonstrates that direct vision supervision enhances the model's fine-grained visual understanding capability rather than merely improving task performance.

During DV-SFT training, vision labels supervise visual tokens in a one-to-one manner. To investigate whether this learning pattern can generalize to one-to-many or many-to-one scenarios, we evaluate the model's ability to recognize text in images at different resolutions. As shown in the results in the right part of Figure~\ref{fig:ocr}, DV-SFT consistently outperforms SFT across different resolutions. Although the model learns vision--text correspondences in a one-to-one mode during training, it effectively acquires visual features rather than mechanically learning word prediction.

\subsection{General Visual Capability}
\begin{table}[]
    \centering
    \begin{tabular}{l|cc|cc}
    \toprule
   \multirow{2}{*}{Method} & \multicolumn{2}{c|}{\small{Qwen3-VL-2B}} & \multicolumn{2}{c}{\small{Qwen3-VL-8B}} \\
   \cline{2-3}\cline{4-5}
     & Acc. & F1  & Acc. & F1\\
    \midrule
    SFT     &  \textbf{81.21} & 80.07 &  87.95 & 88.46\\
    DV-SFT     & 80.88 & \textbf{80.12 } & \textbf{88.08} & \textbf{88.47}\\
    \bottomrule
    \end{tabular}
    \caption{The results on MME.
    }
    \label{tab:mme}
\end{table}

\begin{figure}
    \centering
    \includegraphics[width=\linewidth]{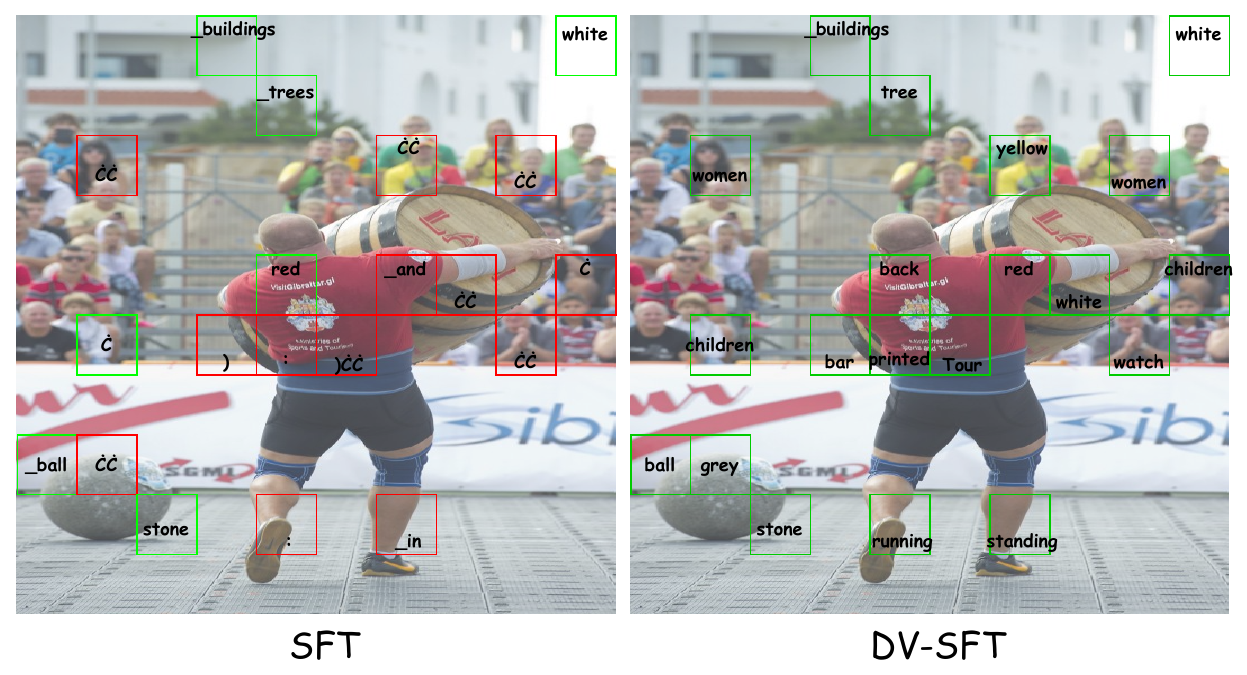}
    \caption{
    A case study of visual logits in a general visual scene.
    The image is resized to dimensions of 10 × 10 visual tokens.
    Each box represents a visual token, with the word inside indicating its Top-1 logit. Green boxes denote matches with the visual content; red boxes indicate mismatches.
    }
    \label{fig:vistok_case}
\end{figure}

We leverage the straightforward association between vision and text in OCR-related scenarios to construct supervisory signals for visual tokens. Specifically, we use the word within the patch corresponding to each visual token to supervise the model's learning of that token. Extensive experiments demonstrate that vision supervision effectively enhances the model's character recognition ability. However, whether purely text-based vision supervision leads to degradation in visual understanding on non-text scenarios remains an open question.

We evaluate the model's visual capability in general scenarios using MME \citep{mme}, a benchmark comprising 10 visual perception tasks and 4 visual cognition tasks. The experimental results are shown in Table~\ref{tab:mme}.
From the results, we can observe that purely word-based vision supervision does not degrade the model's visual capabilities in general scenarios; rather, it slightly improves the model's overall visual performance. This suggests that although vision supervision constructed around OCR scenarios cannot fully annotate visual information, this form of direct vision supervision facilitates the development of fine-grained visual understanding mechanisms in the model.
As can be seen from Figure~\ref{fig:vistok_case}, although the model learns visual features under word-based vision supervision, this direct supervision enables the model to develop more concrete semantic understanding of visual tokens, and this visual advantage generalizes effectively to general scenarios.

In summary, although training the model on OCR-related scenarios yields effective performance improvements on general vision tasks, vision supervision can effectively enhance the model's semantic understanding of visual tokens, and this semantic understanding generalizes from text to general visual scenarios. This phenomenon further validates the effectiveness and necessity of vision supervision.

\subsection{Ablation Study}
\begin{figure}
    \centering
    \includegraphics[width=0.9\linewidth]{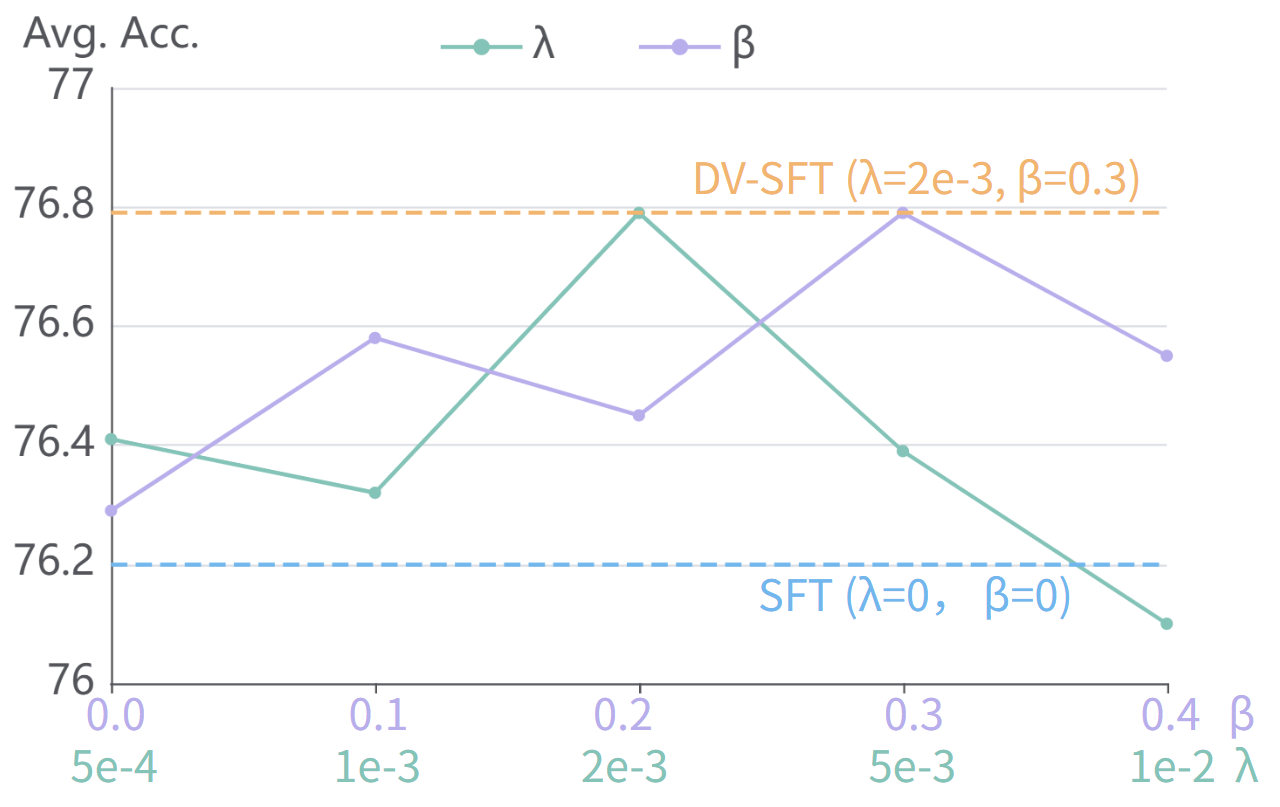}
    \caption{Ablation studies on the validation set of DocVQA and InfographicVQA.
    \emph{Avg. Acc.} denotes average accuracy.
    }
    \label{fig:ablation}
\end{figure}

We conduct ablation experiments on the two key mechanisms in DV-SFT: \emph{Vision Supervision} ($\lambda$ in Eq.~\ref{Eq_lambda}) and \emph{Visual Smoothing} ($\beta$ in Eq.~\ref{Eq_vsm}). The Results are shown in Figure~\ref{fig:ablation}.

\paragraph{Vision Supervision.}
The experimental results show that incorporating only weak vision supervision on top of SFT can yield stable performance improvements for the model. However, we observe that excessively large weights for vision supervision lead to negative effects.
We attribute this phenomenon to three factors:
(1) \emph{Label imbalance}. As shown in the dataset statistics in Table~\ref{tab:data_info}, the number of vision labels is several tens of times that of text labels. Although label normalization alleviates the quantitative imbalance to some extent, a larger amount of supervisory signals can still lead to training imbalance.
(2) \emph{Initial gradient}. We compute the validation loss on 1,024 training samples for the 2B model. The average loss for vision supervision is 24.30, while that for text supervision is 0.98. Since vision labels and text labels follow exactly the same training procedure, the large discrepancy in initial loss causes the gradient to be heavily biased toward learning visual tokens.
(3) \emph{Task objective}. The ultimate goal of MLLMs is to generate responses that meet task requirements, whereas the role of vision supervision is to assist the model in understanding visual information.
In summary, incorporating vision supervision on top of text supervision can effectively enhance the model's visual understanding capability, but the weight of vision supervision should be properly balanced to prevent degradation of the model's generation ability.

\paragraph{Vision Smoothing.}
We can observe that without vision smoothing, the advantage of DV-SFT over SFT is very limited, which is due to the information interaction across visual tokens. Although the forward propagation in MLLMs follows a strictly unidirectional mask, visual tokens first pass through the vision encoder to extract features before being fed into the MLLM. During this process, bidirectional information flow occurs, such that each visual token contains not only the visual information of its corresponding patch but also the context of the entire visual input. Consequently, having each visual token learn only its own label restricts its information representation. In contrast, our proposed visual smoothing method better accommodates the global nature of visual tokens and can more effectively supervise the learning of visual tokens.

In summary, the experimental results validate the necessity of balancing vision loss and text loss, as well as the positive effect of vision smoothing on vision supervision.

\section{Conclusion}
In this paper, we propose DV-SFT, a method motivated by the semantic alignment observed in visual logits that constructs explicit token-level supervision for visual tokens. By exploiting the straightforward vision--text correspondence in OCR-related scenarios, we construct fine-grained vision labels through two complementary pipelines: an image-to-label approach that extracts words from images using OCR engines, and a label-to-image approach that synthesizes images from word labels using drawing tools. To further address the contextualization of visual tokens caused by the bidirectional attention in the vision encoder, we introduced Vision Smoothing, which optimizes the learning of vision labels. Experimental results on both in-domain and out-of-domain benchmarks demonstrate that DV-SFT consistently outperforms standard supervised fine-tuning, and further analyses confirm that direct vision supervision effectively enhances fine-grained visual understanding and improves multimodal alignment efficiency.

In the future, we plan to extend direct vision supervision to broader visual domains beyond OCR.

\section*{Limitations}

We proposed DV-SFT, which enhances the model's visual capability through direct vision supervision.
Although the importance of direct vision supervision has been thoroughly validated, there remains room for improvement in its specific implementation, particularly in the following aspects:

\paragraph{OCR Scenarios.}
We exploited the natural correspondence between vision and text in OCR tasks to construct labels for visual tokens. However, this approach to constructing vision labels may not be easily generalizable to broader visual scenes.
In general visual scenes, visual tokens often carry more complex and diverse information, making it highly challenging to construct sufficiently matched vision labels.

\paragraph{Single Vision Label.}
In OCR tasks, visual tokens and text tend to exhibit a one-to-one correspondence, which facilitates the construction of vision labels. In general visual scenes, however, a single visual token often contains highly rich visual information, and may need multiple labels to represent the visual information within a patch.

\paragraph{Vision Coverage.}
Although our method achieves substantial vision coverage, some visual tokens still remain unlabeled. In future work, we will explore how to improve the construction of vision labels to achieve higher-quality vision supervision.

Despite the above limitations, the experiments fully validate the effectiveness and superiority of direct vision supervision and offer a new perspective for future MLLM training paradigms.

\section*{Ethical Considerations}
This work does not introduce any new datasets and therefore raises no privacy or personally identifiable information concerns. Our method relies solely on existing, publicly available tools for label construction. Consequently, DV-SFT poses negligible unique ethical risks beyond those already associated with general MLLM training and deployment.

\bibliography{custom}

@inproceedings{llava,
  title={Visual instruction tuning},
  author={Liu, Haotian and Li, Chunyuan and Wu, Qingyang and Lee, Yong Jae},
  booktitle={NIPS},
  volume={36},
  pages={34892--34916},
  year={2023}
}

@article{qwen3-vl,
  title={Qwen3-vl technical report},
  author={Bai, Shuai and Cai, Yuxuan and Chen, Ruizhe and Chen, Keqin and Chen, Xionghui and Cheng, Zesen and Deng, Lianghao and Ding, Wei and Gao, Chang and Ge, Chunjiang and others},
  journal={arXiv preprint arXiv:2511.21631},
  year={2025}
}

@inproceedings{BLIP,
  title={Instructblip: Towards general-purpose vision-language models with instruction tuning},
  author={Dai, Wenliang and Li, Junnan and Li, Dongxu and Tiong, Anthony and Zhao, Junqi and Wang, Weisheng and Li, Boyang and Fung, Pascale N and Hoi, Steven},
  booktitle={NIPS},
  pages={49250--49267},
  year={2023}
}

@inproceedings{SigLIP,
  title={Sigmoid loss for language image pre-training},
  author={Zhai, Xiaohua and Mustafa, Basil and Kolesnikov, Alexander and Beyer, Lucas},
  booktitle={CVPR},
  pages={11975--11986},
  year={2023}
}

@inproceedings{ViT,
  title={An image is worth 16x16 words: Transformers for image recognition at scale},
  author={Dosovitskiy, Alexey and Beyer, Lucas and Kolesnikov, Alexander and Weissenborn, Dirk and Zhai, Xiaohua and Unterthiner, Thomas and Dehghani, Mostafa and Minderer, Matthias and Heigold, Georg and Gelly, Sylvain and others},
  booktitle={ICLR},
  year={2021}
}

@inproceedings{
DualToken,
title={Dualtoken: Towards unifying visual understanding and generation with dual visual vocabularies},
author={Wei Song and Yuran Wang and Zijia Song and Yadong Li and Zenan Zhou and Long Chen and Xu Jhua and Jiaqi Wang and Kaicheng Yu},
booktitle={ICLR},
year={2026},
url={https://openreview.net/forum?id=BpgCOFefcE}
}

@inproceedings{
DyME,
title={Empowering small {vlm}s to think with dynamic memorization and exploration},
author={Jiazhen Liu and Yuchuan Deng and Long Chen},
booktitle={ICLR},
year={2026},
url={https://openreview.net/forum?id=qFDju6ahkk}
}

@inproceedings{
Vision-SR1,
title={Vision-{sr}1: Self-rewarding vision-language model via reasoning decomposition and multi-reward policy optimization},
author={Zongxia Li and Wenhao Yu and Chengsong Huang and Zhenwen Liang and Rui Liu and Fuxiao Liu and Jingxi Chen and Dian Yu and Jordan Lee Boyd-Graber and Haitao Mi and Dong Yu},
booktitle={ICLR},
year={2026},
url={https://openreview.net/forum?id=C1M4ETatgM}
}

@article{ASVR,
  title={Autoregressive semantic visual reconstruction helps vlms understand better},
  author={Wang, Dianyi and Song, Wei and Wang, Yikun and Wang, Siyuan and Yu, Kaicheng and Wei, Zhongyu and Wang, Jiaqi},
  journal={arXiv preprint arXiv:2506.09040},
  year={2025}
}

@inproceedings{BASIC,
  title={Basic: Boosting visual alignment with intrinsic refined embeddings in multimodal large language models},
  author={Tang, Jianting and Wang, Yubo and Cao, Haoyu and Xu, Linli},
  booktitle={ICCV},
  pages={20582--20592},
  year={2025}
}

@inproceedings{LaVer,
  title={Unleashing the intrinsic visual representation capability of multimodal large language models},
  author={Li, Hengzhuang and Zhang, Xinsong and Peng, Qiming and Luo, Bin and Hu, Han and Jiang, Dengyang and Ye, Han-Jia and Zhang, Teng and Jin, Hai},
  booktitle={CVPR},
  pages={11975--11986},
  year={2026}
}

@inproceedings{PatchAligned,
  title={Analyzing fine-grained alignment and enhancing vision understanding in multimodal language models},
  author={Jiang, Jiachen and Zhou, Jinxin and Peng, Bo and Ning, Xia and Zhu, Zhihui},
  booktitle={NIPS},
  volume={38},
  pages={874--899},
  year={2025}
}

@inproceedings{ROSS,
  title={Reconstructive visual instruction tuning},
  author={Wang, Haochen and Zheng, Anlin and Zhao, Yucheng and Wang, Tiancai and Ge, Zheng and Zhang, Xiangyu and Zhang, Zhaoxiang},
  booktitle={ICLR},
  volume={2025},
  pages={14374--14399},
  year={2025}
}

@article{VS,
  title={Mitigating hallucination in visual language models with visual supervision},
  author={Chen, Zhiyang and Zhu, Yousong and Zhan, Yufei and Li, Zhaowen and Zhao, Chaoyang and Wang, Jinqiao and Tang, Ming},
  journal={arXiv preprint arXiv:2311.16479},
  year={2023}
}

@article{emu3,
  title={Emu3: Next-token prediction is all you need},
  author={Wang, Xinlong and Zhang, Xiaosong and Luo, Zhengxiong and Sun, Quan and Cui, Yufeng and Wang, Jinsheng and Zhang, Fan and Wang, Yueze and Li, Zhen and Yu, Qiying and others},
  journal={arXiv preprint arXiv:2409.18869},
  year={2024}
}

@inproceedings{EchoGen,
  title={Echogen: Cycle-consistent learning for unified layout-image generation and understanding},
  author={Zou, Kai and Liu, Hongbo and Zheng, Dian and Gao, Jianxiong and Zhao, Zhiwei and Liu, Bin},
  booktitle={AAAI},
  volume={40},
  number={16},
  pages={14068--14076},
  year={2026}
}

@inproceedings{DS-VLM,
  title={Ds-vlm: Diffusion supervision vision language model},
  author={Sun, Zhen and Shen, Yunhang and Li, Jie and Sun, Xing and Dai, Pingyang and Cao, Liujuan and Ji, Rongrong},
  booktitle={ICML},
  year={2025}
}

@article{VGA,
  title={Tell model where to look: Mitigating hallucinations in mllms by vision-guided attention},
  author={Zhao, Jianfei and Zhang, Feng and Sun, Xin and Feng, Chong and Tan, Zhixing},
  journal={arXiv preprint arXiv:2511.20032},
  year={2025}
}

@inproceedings{IC,
  title={Interpreting and editing vision-language representations to mitigate hallucinations},
  author={Jiang, Nick and Kachinthaya, Anish and Petryk, Suzanne and Gandelsman, Yossi},
  booktitle={ICLR},
  volume={2025},
  pages={63582--63605},
  year={2025}
}

@inproceedings{ocrvqa,
  title={Ocr-vqa: Visual question answering by reading text in images},
  author={Mishra, Anand and Shekhar, Shashank and Singh, Ajeet Kumar and Chakraborty, Anirban},
  booktitle={International conference on document analysis and recognition},
  pages={947--952},
  year={2019},
  organization={IEEE}
}

@inproceedings{docvqa,
  title={Docvqa: A dataset for vqa on document images},
  author={Mathew, Minesh and Karatzas, Dimosthenis and Jawahar, CV},
  booktitle={Proceedings of the IEEE/CVF winter conference on applications of computer vision},
  pages={2200--2209},
  year={2021}
}

@inproceedings{infovqa,
  title={Infographicvqa},
  author={Mathew, Minesh and Bagal, Viraj and Tito, Rub{\`e}n and Karatzas, Dimosthenis and Valveny, Ernest and Jawahar, CV},
  booktitle={Proceedings of the IEEE/CVF Winter Conference on Applications of Computer Vision},
  pages={1697--1706},
  year={2022}
}

@inproceedings{squad,
  title={Squad: 100,000+ questions for machine comprehension of text},
  author={Rajpurkar, Pranav and Zhang, Jian and Lopyrev, Konstantin and Liang, Percy},
  booktitle={EMNLP},
  pages={2383--2392},
  year={2016}
}

@article{ocrbench,
  title={Ocrbench: On the hidden mystery of ocr in large multimodal models},
  author={Liu, Yuliang and Li, Zhang and Huang, Mingxin and Yang, Biao and Yu, Wenwen and Li, Chunyuan and Yin, Xu-Cheng and Liu, Cheng-Lin and Jin, Lianwen and Bai, Xiang},
  journal={Science China Information Sciences},
  volume={67},
  number={12},
  pages={220102},
  year={2024},
  publisher={Springer}
}

@inproceedings{chartqa,
  title={Chartqa: A benchmark for question answering about charts with visual and logical reasoning},
  author={Masry, Ahmed and Do, Xuan Long and Tan, Jia Qing and Joty, Shafiq and Hoque, Enamul},
  booktitle={Findings of ACL},
  pages={2263--2279},
  year={2022}
}

@article{HalluSurvey,
  title={A survey on hallucination in large vision-language models},
  author={Liu, Hanchao and Xue, Wenyuan and Chen, Yifei and Chen, Dapeng and Zhao, Xiutian and Wang, Ke and Hou, Liping and Li, Rongjun and Peng, Wei},
  journal={arXiv preprint arXiv:2402.00253},
  year={2024}
}

@article{CICD,
  title={Cross-image contrastive decoding: Precise, lossless suppression of language priors in large vision-language models},
  author={Zhao, Jianfei and Zhang, Feng and Sun, Xin and Kong, Lingxing and Tan, Zhixing and Feng, Chong},
  journal={arXiv preprint arXiv:2505.10634},
  year={2025}
}

@inproceedings{textvqa,
    title={Towards vqa models that can read},
    author={Singh, Amanpreet and Natarjan, Vivek and Shah, Meet and Jiang, Yu and Chen, Xinlei and Batra, Dhruv and Parikh, Devi and Rohrbach, Marcus},
    booktitle={CVPR},
    pages={8317-8326},
    year={2019}
}

@inproceedings{xsum,
  title={Don’t give me the details, just the summary! topic-aware convolutional neural networks for extreme summarization},
  author={Narayan, Shashi and Cohen, Shay B and Lapata, Mirella},
  booktitle={EMNLP},
  pages={1797--1807},
  year={2018}
}

@inproceedings{mme,
  title={Mme: A comprehensive evaluation benchmark for multimodal large language models},
  author={Fu, Chaoyou and Chen, Peixian and Shen, Yunhang and Qin, Yulei and Zhang, Mengdan and Lin, Xu and Yang, Jinrui and Zheng, Xiawu and Li, Ke and Sun, Xing and others},
  booktitle={NIPS},
  year={2026}
}

@article{PaddleOCR,
  title={Paddleocr 3.0 technical report},
  author={Cui, Cheng and Sun, Ting and Lin, Manhui and Gao, Tingquan and Zhang, Yubo and Liu, Jiaxuan and Wang, Xueqing and Zhang, Zelun and Zhou, Changda and Liu, Hongen and others},
  journal={arXiv preprint arXiv:2507.05595},
  year={2025}
}

\clearpage

\appendix
\section{Detailed Experimental Setup}\label{sec:detailed setup}

\subsection{Datasets}
\begin{figure*}
    \centering
    \includegraphics[width=0.6\linewidth]{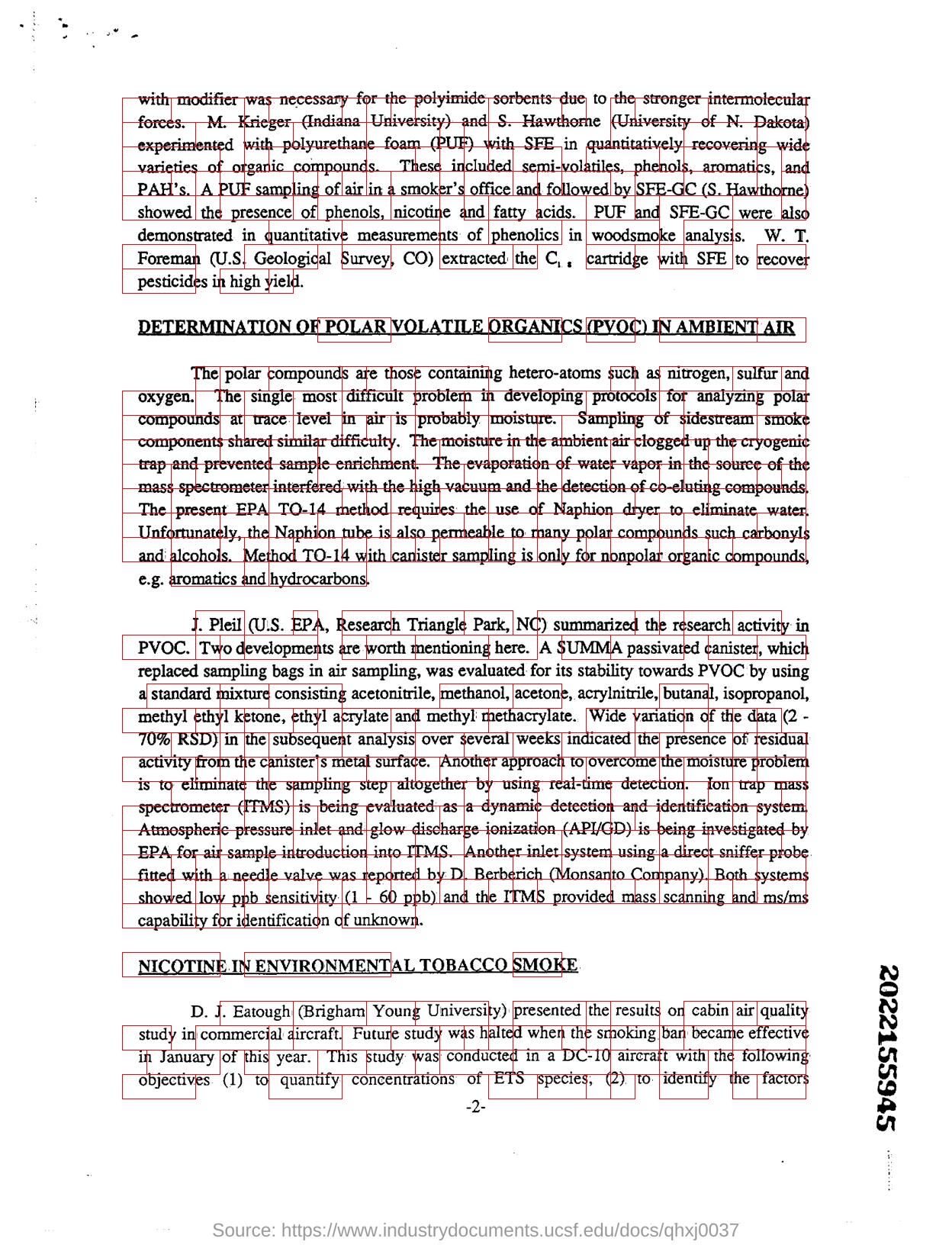}
    \caption{
    An example of constructing vision labels by the image-to-label approach. Each rectangle represents a visual token span that has been successfully aligned with a word.
    }
    \label{fig:docvqa_case}
\end{figure*}

We employ DocVQA \citep{docvqa}, InfographicVQA \citep{infovqa}, and SQuAD \citep{squad} to build training data.
Both DocVQA and InfographicVQA are multimodal datasets, and we construct vision labels using the image-to-label method.
We use PaddleOCR \citep{PaddleOCR} to perform word-level OCR, which specifically includes a text detection model (PP-OCRv5\_server\_det) and a text recognition model (PP-OCRv5\_server\_rec).
An example of image-to-label is shown in Figure~\ref{fig:docvqa_case}.

\begin{figure*}
    \centering
    \includegraphics[width=0.8\linewidth]{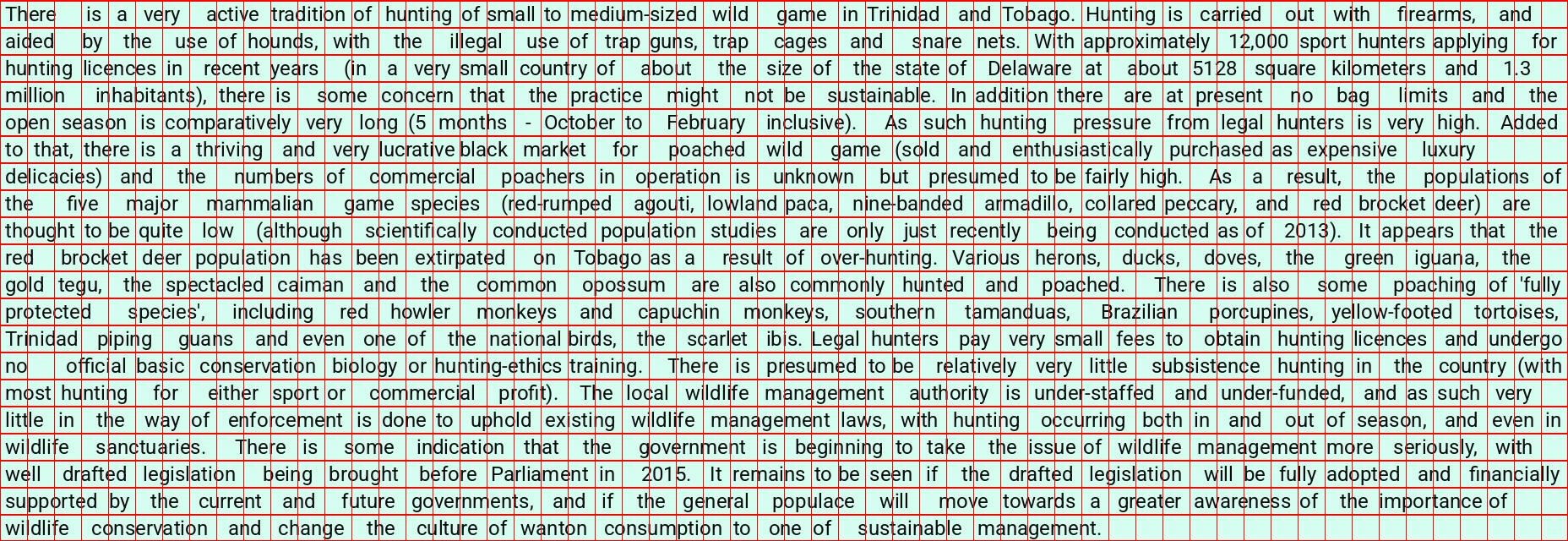}
    \caption{
    An example of constructing vision labels by the label-to-image approach. The grid represents the vision tokenization result, with each box corresponding to a visual token.
    }
    \label{fig:squad_case}
\end{figure*}

SQuAD is a unimodal document understanding dataset with extractive answers, which aligns well with the model's need for accurate visual perception.
We convert this dataset into a multimodal task through the label-to-image approach.
We randomly select one dark and one light color for rendering the background and text, respectively, to enhance training diversity.
An example of label-to-image is shown in Figure~\ref{fig:squad_case}.
To strengthen the model's reliance on visual information, we filter out question-answer pairs in SQuAD that exhibit weak relevance to the document. We also remove answers with low fluency.
We achieve this using the perplexity of a unimodal LLM, Qwen3-0.6B. Specifically, each training sample satisfies the following conditions: $\mathrm{PPL}(A\mid Q) > 100$ and $\mathrm{PPL}(A\mid D,Q) < 10$.

\begin{figure}
    \centering
    \includegraphics[width=\linewidth]{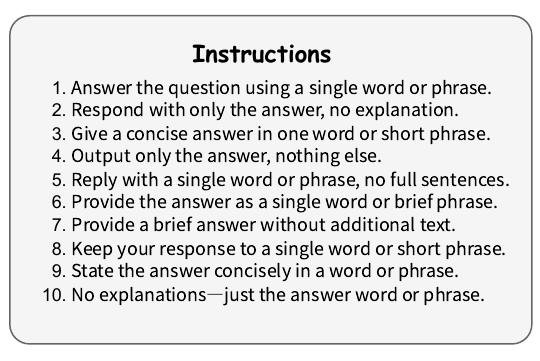}
    \caption{
    Format-control instructions.
    }
    \label{fig:instruction}
\end{figure}

To ensure the verifiability of the model's responses, we randomly incorporate diverse format-control instructions into the training data. The detailed instructions are shown in Figure~\ref{fig:instruction}.

\subsection{Benchmarks}
For in-domain data, we use the test sets of DocVQA and InfographicVQA, as well as the validation set of SQuAD. For out-of-domain data, we use the test set of ChartQA \citep{chartqa}, OCRBench \citep{ocrbench}, the test set of OCRVQA \citep{ocrvqa}, and the validation set of TextVQA \citep{textvqa}.
For OCRVQA, we randomly select 3,000 test samples and ensure that the ground truth of each sample is visible in the image.
Results for DocVQA and InfographicVQA are obtained through online evaluation \footnote{\url{https://rrc.cvc.uab.es/}}, while the remaining benchmarks are evaluated by ground truth.

During testing, we use a uniform format-control instruction: ``Answer the question using a single word or phrase.''

\begin{table}[!ht]
    \centering
    \begin{tabular}{lc}
    \toprule
    \multicolumn{2}{c}{Qwen3-VL-2B / Qwen3-1.7B}  \\
    \midrule
     Min Pixels    &  32*32*64  \\
     Max Pixels    &  32*32*2048  \\
    Global Batch Size  &  64  \\
    Learning Rate    &  5e-6 \\
    Warmup Ratio    &  0.03 \\
    Max Grad Norm    &  1 \\
    LR Scheduler     &  cosine \\
    Weight Decay  & 0 \\
    Model Max Length     &  8192 \\
    \midrule
    \multicolumn{2}{c}{Additional Settings for Qwen3-VL-8B}  \\
    \midrule
    Learning Rate    &  2e-5 \\
    LoRA Rank     &  64  \\
    LoRA Alpha  &  128 \\
    LoRA Dropout & 0 \\
    \bottomrule
    \end{tabular}
    \caption{Detailed training settings.}
    \label{tab:train}
\end{table}

\subsection{Baselines}
We choose the standard SFT method as our baseline, which trains the model using only text labels.
In addition, we reproduce BASIC \citep{BASIC} method. This approach constructs supervisory signals from the hidden states of visual tokens in the early-to-middle layers of the model and trains the projector and the first layer of the model to learn visual features through representation supervision.
To ensure compatibility with FlashAttention, we disable the attention-distribution-based loss normalization in BASIC and adopt uniform normalization instead. 
All hyperparameters follow the settings in the original paper.

\subsection{Implementation Details}
We select Qwen3-VL-2B/8B-Instruct and Qwen3-1.7B as the base models.
For Qwen3-VL-2B/8B-Instruct, we randomly select one QA pair (two for InfographicVQA) per image to avoid duplicate vision labels. For Qwen3-1.7B, we load the visual encoder from Qwen3-VL-2B and use all training data. The visual encoder parameters are frozen in all training settings. Qwen3-VL-2B and Qwen3-1.7B are fully fine-tuned, while Qwen3-VL-8B is fine-tuned using LoRA.
All models are trained for one epoch, and the final checkpoint is selected as the final model.
The detailed experimental settings are shown in Table~\ref{tab:train}.

We set $\beta=0.3$ in Eq. \ref{Eq_vsm} and $\lambda=2\text{e-}3$ in Eq. \ref{Eq_lambda}. We employ greedy decoding during inference with the default setting.
All experiments were conducted on two NVIDIA A100 80G GPUs.

When constructing the Non-Contextual OCR task, we randomly generate unordered word sequences of 200–500 characters in length based on word frequency to create OCR images. The average ground-truth length for the Contextual OCR task is 550.22, while that for the Non-Contextual OCR task is 348.63.

\end{document}